\documentclass[journal]{IEEEtran}

\usepackage{color}

\usepackage{amsmath}
\usepackage{amssymb} 
\usepackage{graphicx}

\usepackage{tikz}
\usetikzlibrary{matrix,chains,positioning,decorations.pathreplacing,arrows}
\usetikzlibrary{fit}
\usepackage{adjustbox}
\usepackage{subfigure}

\DeclareMathOperator \real{\mathbb{R}}

\DeclareMathOperator*{\Ber}{Ber}
\DeclareMathOperator*{\sig}{sig}
\DeclareMathOperator*{\sigh}{sigh}
\newcommand{\Ni}{({\em i})~}
\newcommand{\Nii}{({\em ii})~}
\newcommand{\Niii}{({\em iii})~}
\newcommand{\Niv}{({\em iv})~}

\usepackage{cite}
\usepackage{algorithmic}
\usepackage{caption}

\begin{document}

\title{Impact of Physical Activity on Sleep: \\
A Deep Learning Based Exploration}
%
%
%

\author{Aarti Sathyanarayana, ~\IEEEmembership{Member,~IEEE,}
		Shafiq Joty, ~\IEEEmembership{Member,~IEEE,}
        Luis Fernandez-Luque, ~\IEEEmembership{Member,~IEEE,}
        Ferda Ofli, ~\IEEEmembership{Member,~IEEE,}
        Jaideep Srivastava, ~\IEEEmembership{Fellow,~IEEE,}
        Ahmed Elmagarmid, ~\IEEEmembership{Fellow,~IEEE,}
        Shahrad~Taheri, ~Teresa Arora 
\thanks{A. Sathyanarayana and J. Srivastava are with the Department of Computer Science at the University of Minnesota-Twin Cities.}
\thanks{A. Sathyanarayana, S. Joty, L. Fernandez-Luque, F. Ofli, J. Srivastava, A. Elmagarmid are researchers employed by Qatar Computing Research Institute, Doha, Qatar}
\thanks{S. Taheri and T. Arora are clinical researchers with Weill Cornell Medical College - Qatar. }
\thanks{This work was supported by WCMC-Q and QCRI under IRB 14-00104. }
}

\markboth{ }%
{Sathyanarayana \MakeLowercase{\textit{et al.}}: Impact of Physical Activity on Sleep: A Deep Learning Based Exploration}
%


\maketitle

\begin{abstract}
The importance of sleep is paramount for maintaining physical, emotional and mental wellbeing. Though the relationship between sleep and physical activity is known to be important, it is not yet fully understood. The explosion in popularity of actigraphy and wearable devices, provides a unique opportunity to understand this relationship. Leveraging this information source requires new tools to be developed to facilitate data-driven research for sleep and activity patient-recommendations.

In this paper we explore the use of deep learning to build sleep quality prediction models based on actigraphy data. We first use deep learning as a pure model building device by performing human activity recognition (HAR) on raw sensor data, and using deep learning to build sleep prediction models. We compare the deep learning models with those build using classical approaches, i.e. logistic regression, support vector machines, random forest and adaboost. Secondly, we employ the advantage of deep learning with its ability to handle high dimensional datasets. We explore several deep learning models on the raw wearable sensor output without performing HAR or any other feature extraction.

Our results show that using a convolutional neural network on the raw wearables output  improves the predictive value of sleep quality from physical activity, by an additional 8\% compared to state-of-the-art non-deep learning approaches \cite{arXivRAHAR}, which itself shows a 15\% improvement over current practice \cite{TeresaConvo}. Moreover, utilizing deep learning on raw data eliminates the need for data pre-processing and simplifies the overall workflow to analyze actigraphy data for sleep and physical activity research. 
 
\end{abstract}


%
\IEEEpeerreviewmaketitle

\section{Introduction}
%
%
%
%
\IEEEPARstart{T}{he} importance of sleep is paramount to health and performance. Poor sleep habits impede physical, emotional and mental wellbeing \cite{TeresaConvo, strine2005associations,colten2006sleep}. Insufficient sleep can lead to a multitude of health complications such as insulin resistance \cite{knutson2006role, Gottlieb2005, nilsson2004incidence}, high blood pressure \cite{TIMEarticle}, cardiovascular disease \cite{kasasbeh2006inflammatory, meier2004effect}, a compromised immune or metabolic system \cite{cohen2009,opp2003neural}, mood disorders (such as depression or anxiety)  \cite{peterman2015anxiety, murphy2015sleep}, and decreased cognitive function for memory and judgement \cite{williamson2000moderate, van2003cumulative, ellenbogen2006role}. After decades of exploration, research is still under way to uncover the long-term consequences of inadequate sleep and sleep disorders. 

Diagnostic sleep studies (called polysomnography) are performed daily in thousands of hospitals to diagnose major sleep disorders such as obstructive sleep apnea, restless leg syndrome, insomnia and more \cite{hirshkowitz2015history}. Current processes are cumbersome and require manual interpretation by clinicians. Moreover, patient referral for polysomnography (PSG) is based on self-reported needs; and hence exacerbation often precedes diagnosis. 

Sleep researchers work towards improving PSG tests to better understand the overall impact of sleep on an individual's quality of life and well-being. Formally funded, large-cohort clinical research studies have been a primary resource for data collection to date. However, the recent widespread adoption of wearable devices is now producing a rapidly growing amount of human activity data. Improved analytic tools can leverage this data for clinical practice and advanced health research.

In this paper we explore a deep learning approach to modeling the relationship between sleep and physical activity. We first compare deep learning models with classical models, i.e. logistic regression, support vector machines, random forest, adaboost, on data pre-processed to create a feature space of human activity by an automated human activity recognition (HAR) algorithm. However, deep learning has the advantage that it is robust to high dimensional data. We leverage this characteristic by building models using a range of deep learning methods on raw accelerometer data. This eliminates the need for data pre-processing and feature space construction, and simplifies the overall workflow for clinical practice and sleep researchers.


\subsection{Activity and Sleep Research}

Polysomnography (PSG) involves the use of multiple sensors, such as EEG (electroencephalogram), motion sensors, breathing sensors, SpO2 (oxygen saturation), etc. These combined techniques observe a patient overnight in a clinical setting \cite{hirshkowitz2015history}. A restriction of PSG is its lack of easy portability. Portable solutions would allow the performance of diagnostics in the home of the patient, decreasing costs and inconvenience, and improving the overall evaluation by observing a patient in their natural setting \cite{bruyneel2013real}. Currently, only a few portable sleep study solutions are approved and available on the market. Although many studies have shown that they are effective \cite{hwanghome}, limited recording channels make home studies less accurate for investigating complex sleep disorders \cite{boraziotoward}. 

PSG is usually limited to one overnight observation. It doesn't observe a patient over time and it does not take into consideration the relationship between sleep quality and an individual's daily physical activity. Although the relationship between physical activity and sleep is not yet fully understood \cite{driver2000exercise}, it is thought to have a strong and complex correlation and contribute to multiple lifestyle diseases such as type II diabetes mellitus and obesity \cite{chennaoui2015sleep}. 

In the 1990's, researchers developed a technique called actigraphy to study physical activity and sleep interactions. Actigraphy uses motions sensors in wearable devices to measure human behaviours, such as sleep and physical activity. There are hundreds of quantified self-applications syncing with such devices to help individual's self-monitor their physical fitness and well-being. Actigraphy has become a widely used tool as it has been found to be much more reliable than subjective or self-reported sleep diaries and behaviour logs \cite{arora2013investigation}. This has been especially influential for large cohort studies about physical activity and sleep where PSG is not feasible \cite{lee2014using}. Moreover, actigraphy allows for the continuous longitudinal monitoring of a patient/participant. This is particularly impactful for the study of diseases such as Chronic Obstructive Pulmonary Disease (COPD) where sleep disturbances can be a predictor of exarcebation of the disease \cite{ehsan2013longitudinal}. In this example, it is the long term evolution of the disease that can reveal potential health decline. In such settings, actigraphy is the best and only tool to study sleep and physical activity over long periods. 

\subsection{Research challenges of Actigraphy}
In sleep research, there is a major bottleneck for the analysis of actigraphy data. The process involves sleep experts manually configuring parameters prior to performing analysis. In our previous study \cite{arXivRAHAR} we automated this process by developing a robust automated human activity recognition algorithm (called RAHAR). RAHAR automatically pre-processes accelerometer data into meaningful activity levels. The dimensionality reduction caused by RAHAR's feature construction creates a dataset suitable for effective classical model building, and leads to an improvement of the predictive value for sleep quality by 15\%. 

However, this approach has some limitations. Firstly, RAHAR is an unsupervised algorithm, meaning it does not exploit task labels for feature construction, and thus can be limited in its ability to learn task specific features. Secondly, it is a pre-processing feature-reduction step. RAHAR adds to the evaluation workflow and summarizes the data through aggregation, rather than exploiting its richness. 

An alternative approach that has become popular recently is to use the automatic feature learning capability offered by deep learning methods. Deep learning models have achieved state-of-the-art results in a wide variety of tasks in computer vision, natural language processing and speech recognition. The fact that deep learning models automatically learn abstract feature representations from raw features, while also optimizing on the target prediction tasks, makes them an attractive solution for sleep and health research.


\section{Deep learning with Activity Data}

As explained before, sleep is highly affected by physical activity during the day. Therefore, it is of special importance to consider how deep learning can be used to study activity data acquired from sensors such as actigraphy. 

Over the past decade deep learning has gained significant momentum in areas such as speech recognition, natural language processing and computer vision. This trend has recently expanded into various other areas including human activity recognition from sensor data. In this sub-section, we limit our deep learning review to studies that employ techniques to analyses of physical activity data, and refer the reader to Chen et al.~\cite{LChen:TCMCC12} and Bulling et al.~\cite{ABulling:CSUR14} for more comprehensive reviews of sensor-based human activity recognition literature. Also, Langkvist et al. provided a review of deep learning for time-series modeling \cite{MLangkvist:PatRec14}. 


Early adopters of deep learning for wearable sensor data, focus on adapting the raw input signal to the desired deep learning architecture \cite{MAlsheikh:AAAIW16}. Inspired from speech recognition, Alsheikh et al.~\cite{MAlsheikh:AAAIW16}, for example, used the spectrogram representation to convert one-dimensional accelerometer data into two-dimensional data input of a deep restricted Boltzmann machine (RBM) architecture.

Instead of resizing the data to meet the input format of a conventional convolutional neural network (CNN), Chen and Xue~\cite{YChen:SMC15} suggested resizing the convolutional kernels to deal with the original tri-axial acceleration data with different lengths in order to preserve the temporal information between adjacent acceleration values.

Similarly, Yang et al.~\cite{JYang:IJCAI15} employed a deep CNN architecture to automate feature learning from raw multi-sensor inputs in a systematic way in which the convolution and pooling filters were applied along the temporal dimension for each sensor, and the resulting feature maps for different sensors were unified as a common input for the neural network classifier.

Considering recurrent neural networks (RNN) model the long-term contextual information of temporal sequences well, Du et al.~\cite{YDu:CVPR15} proposed a hierarchical RNN for action recognition from skeletal movement data. Instead of taking the whole skeleton as input, they considered dividing the skeleton into five parts which were then fed into five bidirectional-RNNs. The representations extracted by the subnets were hierarchically fused as the inputs of higher layers. A fully connected layer and a softmax layer were performed on the final representation to classify the actions, achieving state-of-the-art results in skeleton-based action recognition.

Lefebvre et al.~\cite{GLefebvre:LNCS15} developed a gesture recognition method based on bidirectional long short-term RNN (BLSTM-RNN) from raw input data, which outperformed the classical methods such as those based on hidden Markov models (HMM), dynamic time warping (DTW), or support vector machines (SVM), on a common dataset for 14-class gesture classification.

Some of the latest studies in this domain have introduced more sophisticated deep learning architectures. For instance, Neverova et al.~\cite{NNeverova:Access16} proposed an optimized shift-invariant dense convolutional mechanism, and incorporated the discriminatively trained dynamic features in a probabilistic generative framework, taking into account temporal characteristics of the input physical activity data at multiple scales. Their method proved successful at learning human identity from daily motion patterns, potentially allowing for active biometric authentication with mobile inertial sensors.

Inspired by recent advances in speech recognition, Ordonez and Roggen recently proposed an architecture comprising both convolutional and LSTM recurrent neural networks for activity recognition from multimodal wearable sensor data\cite{FOrdonez:Sensors16}. The convolutional layers in this architecture acted as feature extractors and provided abstract representations of the input sensor data in feature maps while the recurrent layers modeled the temporal dynamics of the activation of the feature maps.

Most recently, Hammerla et al.~\cite{NHammerla:IJCAI16} provided the first systematic exploration of the performance of several state-of-the-art deep learning architectures such as deep feed forward, convolutional and recurrent neural networks, on three representative physical activity datasets captured with wearable sensors. In more than 4,000 recognition experiments with randomly sampled model configurations, they investigated the suitability of each model for different tasks in human activity recognition, explored the impact each model's hyper-parameters had on performance using the fANOVA framework, and provided guidelines for those who wanted to apply deep learning in their application scenarios. Over the course of these experiments they discovered that recurrent networks outperformed the state-of-the-art, and that they allowed novel types of real-time applications through sample-by-sample prediction of physical activities.

\section{Preliminaries}
In this section we present the terminology used and a description of how the dataset for our experiments were collected.

\subsection{Sleep Science Terminology}
For our model there are two key types of vocabulary used: those pertaining to time series segmentation, and those for defining sleep quality. 

In a person's activity time series, i.e. the continuous data collected from a wearable device, there are moments when an individual is awake and when they are asleep. The latter is referred to as the \textit{sleep period}. The boundary of the time awake to time asleep is called \textit{sleep onset time}, and the boundary of the time asleep to time awake, is called the \textit{sleep awakening time}. All time extending between a sleep awakening time and the following sleep onset time, is referred to as \textit{awake time}. 

To measure quality of sleep we use a metric called \textit{sleep efficiency}, which is the ratio of total minutes asleep to total minutes in bed. 
\begin{equation}
\begin{split}
\textit{Sleep Efficiency} =  
& \frac{Total Sleep Time}{Total Minutes in Bed} 
\\
\\
& = \frac{length(Sleep Period)-WASO}{length(Sleep Period) + Latency}
\\
\\
\end{split}
\end{equation}

\textit{Total minutes in bed} represents the amount of time that an individual spends sleeping and the amount of time it takes for them to fall asleep, called \textit{latency}. \textit{Total sleep time} represents the amount of time that an individual spends actually sleep. This is calculated by subtracting all moments of wakefulness, \textit{WASO}, from the duration of the sleep period. 
\begin{equation}
\textit{WASO} =  
\sum_{Sleep Period} length(Wakefulness Period)
\end{equation}

\subsection{Data Collection}

The deidentified data used in this study which was collected by Weill Cornell Medical College - Qatar for a research study called Qatar's Ultimate Education for Sleep in Teenagers (QUEST). The study aimed to examine the relationship between sleep and physical activity. The selected cohort was chosen from student volunteers attending two different high schools. The students were provided with an ActiGraph GT3X+ device, to wear on their non-dominant wrist throughout the day and night for seven consecutive days/nights. The dataset used in our experiments contains the actigraphy data from a subset of 92 adolescents over 1 week. 

The ActiGraph GT3X+ is a clinical-grade wearable device that samples a users activity at 30-100 Hertz. The effectivness of this device has been successfully validated against clinical polysomnography \cite{freedson1998calibration}. The device has an accompanying software called ActiLife. This software is used by sleep experts to evaluate a patient's sleep period, and interpret the accelerometer output. We evaluate the performance improvement of our methodology against results from the ActiLife software version 6.  

\section{Methodology}

\subsection{Data Representation}

Our experiments fall into two categories. The first uses output from a human activity recognition algorithm called RAHAR, and the second uses accelerometer triaxial data from the actigraph. In this subsection, we start by describing the details that apply to both data types, and then we discuss the dataset specific components. 

In traditional sleep science, sleep onset time and sleep awakening time are metrics that define the sleep period \cite{berry2012aasm}. We interpret and expand these values for accelerometer data according to Sadeh's actigraphy definitions \cite{sadeh2000sleep}. Sleep onset time is defined as the first minute of 15 continuous minutes of sleep, after a self-reported bedtime, and the sleep awakening time is the last minute of 15 continuous minutes of sleep that is followed by 30 minutes of activity \cite{sadeh2000sleep}. To automate this interpretation directly from the accelerometer output, we reverse engineer the \textit{self-reported bedtime} as the start time of sedentary activity immediately preceding and adjacent to the sleep-period. The sleep period itself is designated by observing epochs that contain no triaxial movement, and then fine tuned with the formal boundaries of sleep onset and sleep awakening. Sleep onset time is the first candidate boundary of the sleep period that is followed by 15 minutes of continuous sleep. Sleep awakening time is the last candidate boundary of 15 minutes of continuous sleep followed by 30 minutes of activity.  

The classification label of each record is based on the sleep quality. Sleep quality is measured by the sleep efficiency equation in section III-A. With the sleep period boundaries defined as mentioned above, WASO can be computed by summing up the total minutes during the sleep period that a person has some movement for more than 5 consecutive minutes. Latency can be computed as the length of time before the sleep onset time, that a person is sedentary. Overall, if sleep efficiency is above 85\%, the sleep is labeled as "Good", otherwise as "Poor". 

The RAHAR algorithm uses the raw accelerometer data to compute the percentage of awake time spent in 1 of 4 activity intensity levels: sedentary, light, moderate, vigorous. Thus the feature set is fixed at 4.  In order to handle class imbalance, synthetic minority oversampling technique  \cite{chawla2002smote} was run on the data. 

The raw accelerometer data was aggregated into minute-by-minute epochs. The sleep period was used to classify the awake time, and it was detached from each time series to prevent auto-correlation. The vertical axis from the accelerometer was used to represent the summary of movement. The raw data was thus one-dimensional.  

\subsection{Deep Learning Models} \label{models}

Let $\mathbf{x_t}\in\real^{D}$ be a vector representing a person's activity measured at time $t$. 
Given a series of such input feature vectors $\mathbf{X} = (\mathbf{x}_1, \cdots, \mathbf{x}_T)$ representing a person's physical activity in an awake time period, the deep neural models first compute compressed representations with multiple levels of abstraction by passing the inputs through one or more non-linear hidden layers. The abstract representations of the raw activity measures are then used in the output layer of the neural network to predict the sleep quality. Formally, the output layer defines a Bernoulli distribution over the sleep quality $y \in $ \{\textit{good, poor}\}:  

\begin{equation}
p(y|\mathbf{X}, \theta)= \Ber(y| \sig(\mathbf{w^T} \phi(\mathbf{X}) + b )) \label{loss}
\end{equation}

\noindent  where $\sig$ refers to the sigmoid function, $\phi(\mathbf{X})$ defines the transformations of the input $\mathbf{X}$ through non-linear hidden layers, $\mathbf{w}$ are the output layer weights and $b$ is a bias term.

We train the models by minimizing the cross-entropy between the predicted distributions $\hat{y}_{n\theta} = p(y_n|\mathbf{X}_n, \theta)$ and the target distributions $y_n$ (i.e., the gold labels).\footnote{Other loss functions (e.g., hinge) yielded similar results.}

\begin{equation}
J(\theta) =  - \sum_{n} y_n \log \hat{y}_{n\theta} + (1-y_n) \log \left(1- \hat{y}_{n\theta} \right) \label{eq:ce}
\end{equation}

Minimizing cross-entropy is same as minimizing the negative log-likelihood (NLL) of the data (or maximizing log-likelihood). Unlike generalized linear models (e.g., logistic regression), the NLL of a deep neural model is a non-convex function of its parameters. Nevertheless, we can find a locally optimal maximum likelihood (ML) or maximum a posterior (MAP) estimate using gradient-based optimization methods. The main difference between the models, as we describe below, is how they compute the abstract representation $\phi(\mathbf{X})$.     

\subsubsection{Multi-Layer Perceptrons} \label{mlp}

\begin{figure}[tb!]
\centering
\includegraphics[scale=1.0]{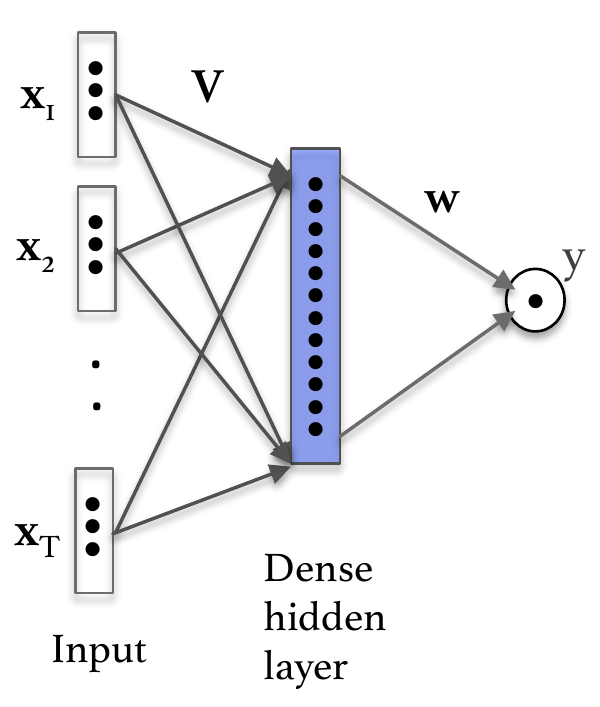}
\label{fig:mlp}
\caption{A multi-layer perceptron with one hidden layer.}
\end{figure}

Multi-Layer Perceptrons (MLP), also known as Feed-forward Neural Networks, are the simplest models in the deep learning family. As shown in fig. 1, the transformation of input $\phi(\mathbf{X})$ in MLP is defined by one or more fully-connected hidden layers of the form  

\begin{equation}
\phi(\mathbf{X}) =  f(V\mathbf{x_{1:T}}) = [f(\mathbf{v}_1^T\mathbf{x_{1:T}}), \cdots , f(\mathbf{v}_N^T\mathbf{x_{1:T}})]
\end{equation}

\noindent where $\mathbf{x_{1:T}}$ is the concatenation of the feature vectors $\mathbf{x}_1, \cdots, \mathbf{x}_T$, $V$ is the weight matrix from the inputs to the hidden units, $f$ is a non-linear activation function (e.g., $\sig, \tanh$) applied element-wise, and $N$ is the number of hidden units. MLP without any hidden layer (i.e., non-linear activations) boils down to a logistic regression (LR) or maximum entropy (MaxEnt) model. The hidden layers give MLP representational power to model complex dependencies between the input and the output.\footnote{In fact, MLP has been shown to be a \emph{universal approximator}, meaning that it can model any suitably smooth function to any desired level of accuracy, given the required hidden units  \cite{Hornik:1991}.} By transforming  a large and diverse set of raw activity measures into a more compressed abstract representation through its hidden layer, MLP can improve the prediction accuracy over linear models like LR. 


\subsubsection{Convolutional Neural Network} \label{cnn}

The fully-connected MLP described above has two main properties: \Ni it composes each higher level feature from the entire input, and \Nii it is \emph{time variant}, meaning it uses separate (non-shared) weight  parameter for each input dimension to predict the overall sleep quality. However, a person's sleep quality may be determined by his activity over certain (local) time periods in the awake time as opposed to the entire awake time, and this can be invariant to specific timings. For example, high intensity exercises or games over certain period of time can lead to good sleep, no matter when the activities are exactly performed in the awake time. Furthermore, each person has his own habit of activities, e.g., some run in the morning while others run in the afternoon. A fully-connected structure would require a lot of data to effectively learn these specific activity patterns, which is rarely the case in health domain. Convolutional neural networks (CNN) address these issues of a fully-connected MLP by having repetitive filters or kernels that are applied to local time slots to compose higher level abstract features. The weights for these filters are shared across time slots. 

As shown in Fig. 2, the hidden layers in a CNN are formed by a sequence of \textit{convolution} and \textit{pooling} operations. A convolution operation involves applying a \emph{filter} $\mathbf{u} \in \real^{L.D}$ to a window of $L$ feature vectors to produce a new feature $h_t$

\begin{equation}
h_t = f(\mathbf{u} . \mathbf{x}_{t:t+L-1})
\end{equation}

\noindent where $\mathbf{x}_{t:t+L-1}$ denotes the concatenation of $L$ input vectors and  $f$ is a non-linear activation function as defined before. We apply this filter to each possible $L$-word window in the sequence $\mathbf{X}$ to generate a \emph{feature map} $\mathbf{h}^i = [h_1, \cdots, h_{T+L-1}]$. We repeat this process $N$ times with $N$ different filters to get $N$ different feature maps $[\mathbf{h}^1, \cdots, \mathbf{h}^N]$. Note that we use a \emph{wide} convolution rather than a \emph{narrow} one, which ensures that the filters reach the entire sentence, including the boundary words \cite{Kalchbrenner14}. This is done by performing \emph{zero-padding}, where out-of-range ($t$$<$$1$ or $t$$>$$T$) vectors are assumed to be zero. 


\begin{figure}[tb!]
\centering
\includegraphics[scale=0.8]{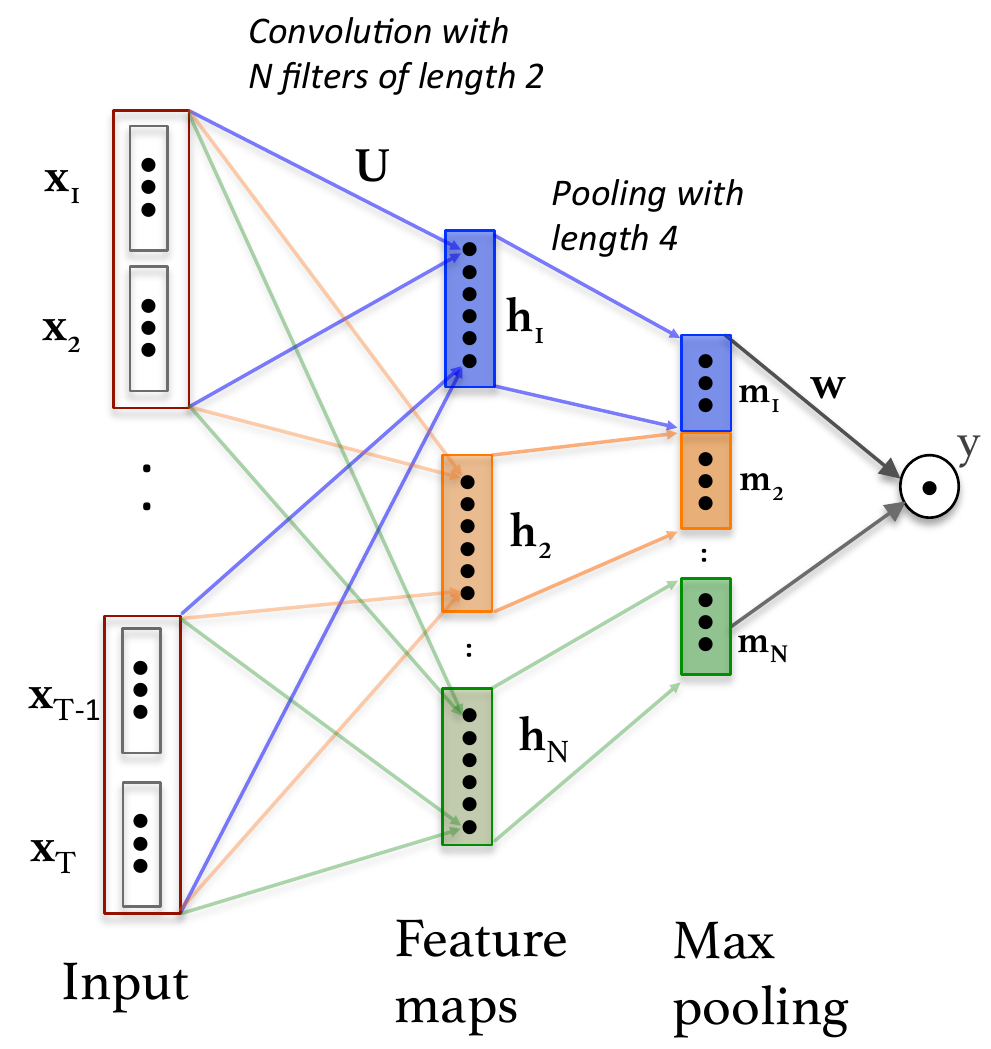}
\label{fig:cnn}
\caption{A convolutional neural network. Each colored box in the second hidden layer represents a feature map, which is obtained by sliding its corresponding filter over the entire input. Pooling is independently done over feature maps.}
\end{figure}

After the convolution, we apply a max-pooling operation\footnote{Other pooling operations are possible. In our experiments, we found max-pooling giving better results than mean-pooling.} to each feature map

\begin{equation}
\mathbf{m} = [\mu_p(\mathbf{h}^1), \cdots, \mu_p(\mathbf{h}^N)] \label{max_pool}
\end{equation}

\noindent where $\mu_p(\mathbf{h}^i)$ refers to the $\max$ operation applied to each window of $p$ features in the feature map $\mathbf{h}^i$. For $p=2$, this pooling gives the same number of features as in the feature map (because of the zero padding). 

Intuitively, the filters compose activity measures in local time slots into higher-level representations in the feature maps, and max-pooling reduces the output dimensionality while keeping the most important aspects from each feature map. Since each convolution-pooling operation is performed independently, the features extracted become invariant in locations, i.e., where they occur in the awake time. This design of CNNs yields fewer parameters than its fully-connected counterpart, therefore, generalizes well for target prediction tasks.  



\subsubsection{Recurrent Neural Network} \label{rnn}

\begin{figure*}[t!]
\centering
\subfigure[]{%
\includegraphics[scale=0.84]{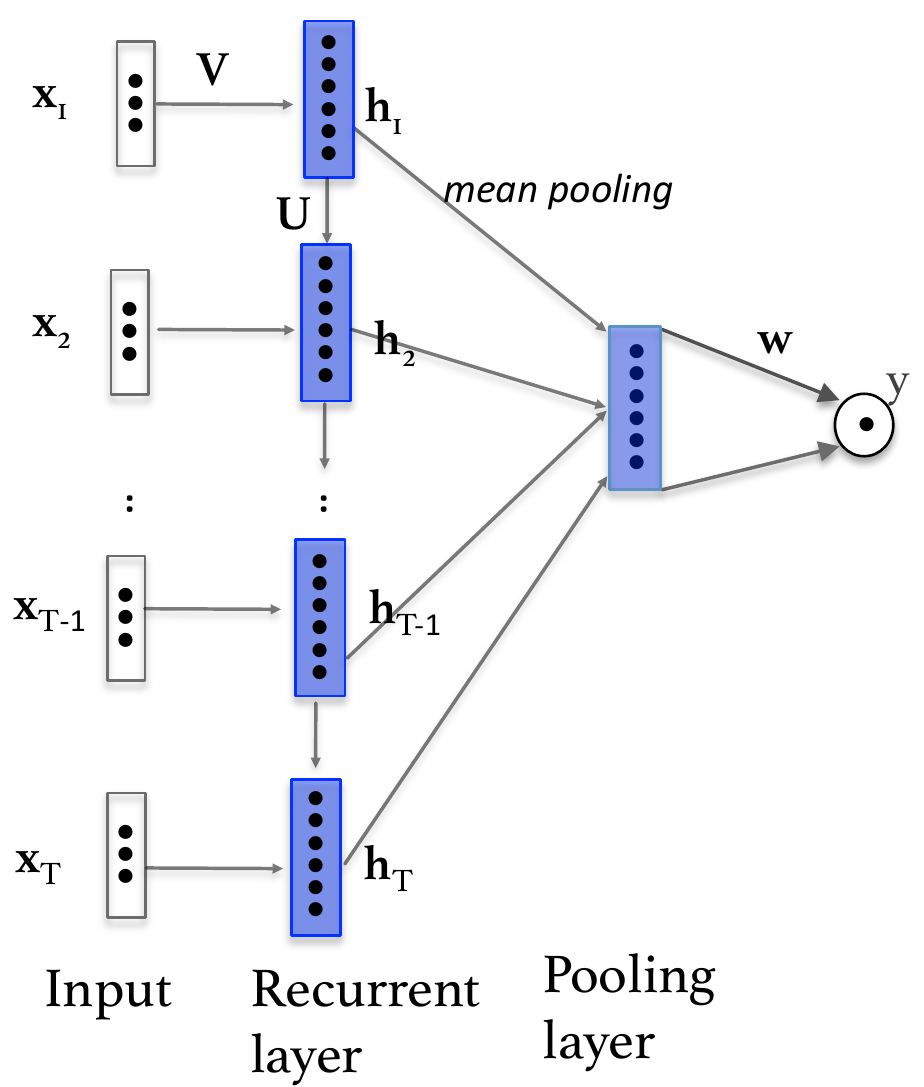}
\label{fig:rnn}}
\quad
\subfigure[]{
\raisebox{20mm}{\includegraphics[scale=0.4]{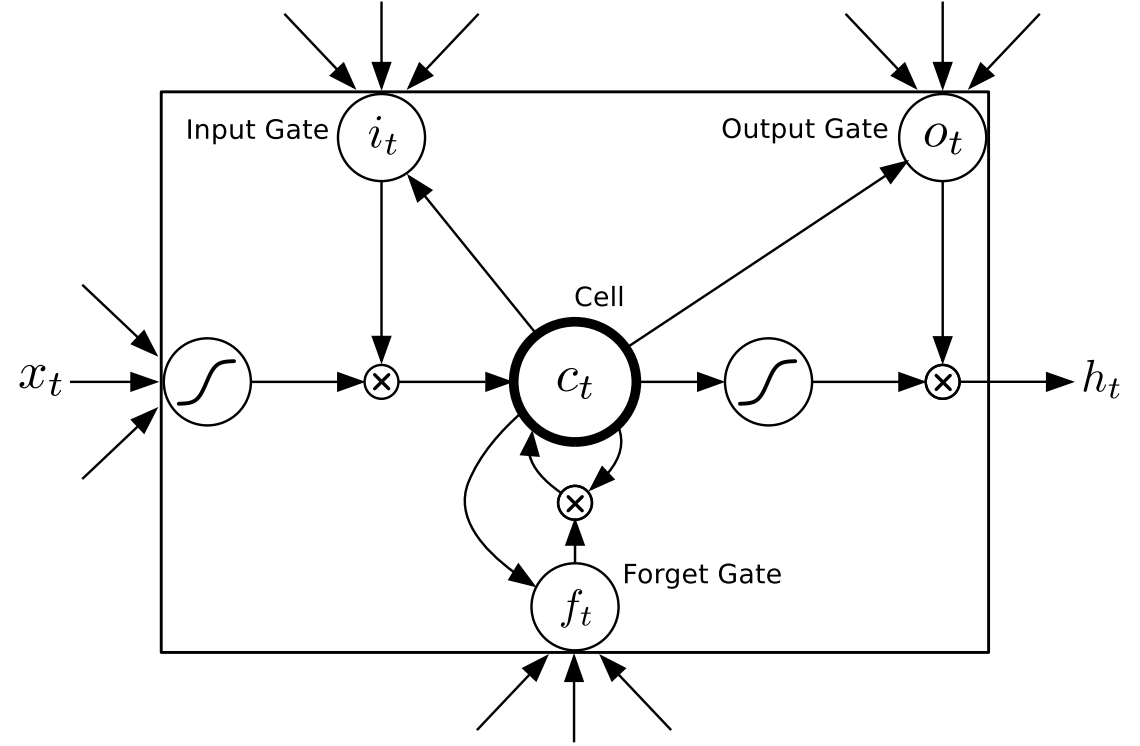}}
\label{fig:lstm}}
\caption{ (a) A recurrent neural network with one recurrent layer, and (b) an LSTM memory block, which represents a hidden unit in an LSTM-RNN.}
\end{figure*}

In CNN, features are considered in a bag-of-words fashion disregarding the order information. The order in which the activities were performed in an awake time could be important. Recurrent neural networks (RNN) compose abstract features by processing activity measures in an awake time sequentially, at each time step combining the current input with the previous hidden state. More specifically, as depicted in Fig. \ref{fig:rnn}, RNN computes the output of the hidden layer $\mathbf{h}_t$ at time $t$ from a nonlinear transformation of the current input $\mathbf{x}_t$ and the output of the previous hidden layer $\mathbf{h}_{t-1}$. More formally,

\begin{equation}
\mathbf{h}_t = f( U \mathbf{h}_{t-1} + V \mathbf{x}_t) \label{rec}
\end{equation}

\noindent where $f$ is a nonlinear activation function as before, and $U$ and $V$ are compositional weight matrices. RNNs create internal states by remembering previous hidden layer, which allows them to exhibit dynamic temporal behavior. We can interpret $\mathbf{h}_t$ as an intermediate representation summarizing the past. The representation for the entire sequence can be obtained by performing a pooling operation (e.g., \emph{mean-pooling, max-pooling}) over the sequence of hidden layers or simply by picking the last hidden layer $\mathbf{h}_T$. In our experiments, we found mean-pooling to be more effective than other methods.     


RNNs are generally trained with the backpropagation through time (BPTT) algorithm, where errors (i.e., gradients) are propagated back through the edges over time. One common issue with BPTT is that as the errors get propagated, they may soon become very small or very large that can lead to undesired values in weight matrices, causing the training to fail. This is known as the \emph{vanishing} and the \emph{exploding} gradients problem \cite{Bengio:1994}. One simple way to overcome this issue is to use a truncated BPTT \cite{Mikolov12} for restricting the backpropagation to only few steps like $4$ or $5$. However, this solution limits the simple RNN to capture long-range dependencies. Below we describe an elegant RNN architecture to address this problem.

\paragraph*{Long Short-Term Memory RNN}
Long short-term memory or LSTM \cite{Hochreiter:1997} is specifically designed to capture long range dependencies in RNNs. The recurrent layer in a standard LSTM is constituted with special units called \emph{memory blocks} (Fig. \ref{fig:rnn} and \ref{fig:lstm}). A memory block is composed of four elements: \Ni a memory cell $c$ (a neuron) with a self-connection, \Nii an input gate $i$ to control the flow of input signal into the neuron, \Niii an output gate $o$ to control the effect of the neuron activation on other neurons, and \Niv a forget gate $f$ to allow the neuron to adaptively reset its current state through the self-connection. The following sequence of equations describe how the memory blocks are updated at every time step $t$: 

\begin{eqnarray}
	\mathbf{i}_t &=& \sigh(U_i\mathbf{h}_{t-1} + V_i\mathbf{x}_t + \mathbf{b}_i)  \label {lstm_first}\\
    \mathbf{f}_t &=& \sigh(U_f\mathbf{h}_{t-1} + V_f\mathbf{x}_t + \mathbf{b}_f) \\
    \mathbf{c}_t &=& \hspace{-0.1cm}  \mathbf{i}_t\odot \tanh(U_c\mathbf{h}_{t-1} + V_c\mathbf{x}_t) + \mathbf{f}_t\odot\mathbf{c}_{t-1} \\
    \mathbf{o}_t &=& \sigh(U_o\mathbf{h}_{t-1} + V_o\mathbf{x}_t + \mathbf{b}_o) \\
    \mathbf{h}_t &=& \mathbf{o}_t \odot \tanh(\mathbf{c}_t) \label {lstm_last}
\end{eqnarray}

\noindent where $U_k$ and $V_k$ are the weight matrices between two consecutive hidden layers, and between the input and the hidden layers, respectively, which are associated with gate $k$ (input, output, forget and cell); and and $\mathbf{b}_k$ is the corresponding bias vector. The symbols $\sigh$ and $\tanh$ denote hard sigmoid and hard tan, respectively, and the symbol $\odot$ denotes a element-wise product of two vectors. LSTM by means of its specifically designed gates (as opposed to simple RNNs) is capable of capturing long range dependencies.




\section{Experiments and Results}

In this section we present the experimental settings and results. The data is partitioned into a $70\%$:$15\%$:$15\%$ split for training, testing and validation, respectively. Although two different types of data were used, the partitions for each contained the same subjects data, i.e. the training set from the human activity recognition data contained the same subject data as the training set from the raw accelerometer data. All reported results are based on model predictions on the test set.

\subsection{Settings for Deep Learning Models}
We train our neural models by optimizing the cross entropy in Eq. \ref{eq:ce} using the gradient-based online learning algorithm RMSprop \cite{Tieleman12}.\footnote{Other adaptive algorithms (ADAM \cite{KingmaB14}, ADADELTA \cite{Zeiler12}) gave similar results.} The learning rate and other parameters ($\rho$ and $\epsilon$) were set to their default values as suggested by the authors. Maximum number of epochs for all models was set to $50$. We use rectified linear units (ReLU) for the activation functions ($f$). To avoid overfitting, we use dropout \cite{Srivastava14a} of hidden units and early stopping based on the loss on the development set.\footnote{Regularization on weights such as $l_1$ and $l_2$ did not improve the results.}  We experimented with $DR \in \{0.0, 0.1, 0.2, 0.3, 0.4, 0.5\}$ for dropout rates and $MB \in \{5, 10, 15, 20\}$ for minibatch sizes. Since the size of the training data is small, the network weights are initialized with zero to start the training with the simplest model.
\paragraph*{MLP Settings}
In our MLP, we experimented with one hidden layer containing $N \in \{ 2, 3, 5, 10, 15, 20\}$ units. Increasing the number of hidden layers worsened the results because of small amount of training data. 
\paragraph*{CNN Settings}
For CNN, we experimented with $N \in \{25, 50, 75, 100, 125, 150\}$ number of filters with filter lengths $FL\in \{2,3,4,5\}$ and pooling lengths $PL \in \{2,3,4,5\}$. 
\paragraph*{RNN Settings}
For simple RNN and LSTM RNN, we experiment with $N \in \{25, 50, 75, 85, 100\}$ units in the recurrent layer and $S \in \{25, 50, 75, 100\}$ time-slots for constructing pseudo-sequences.



\subsection{Human Activity Recognition Data Results}
In our previous study, accelerometer data processed by RAHAR, produced a 15\% improvement over current sleep expert methodology. RAHAR plus random forests performed the best, with linear regression performing similarily well. Due to the dimensionality reduction caused by RAHAR, the dataset contained 4 features. RAHAR also aggregates the data over the time series, which loses the time component. As a result, CNN, RNN, and LSTM are not suitable, and we apply a multi layer perceptron (MLP). 

\subsubsection{Multi Layer Perceptron}
The best configuration for a single layered MLP used a hidden layer size of 15, mini-batch size of 5, and a dropout ratio of 0.3. MLP showed an improvement to the performance of LR, and marginally surpassed the performance of random forest (Table \ref{tab:RAHAR}). 

\begin{figure*}[t!]
\centering
\subfigure[ROC curve for RAHAR data]{%
\includegraphics[width=0.47 \linewidth]{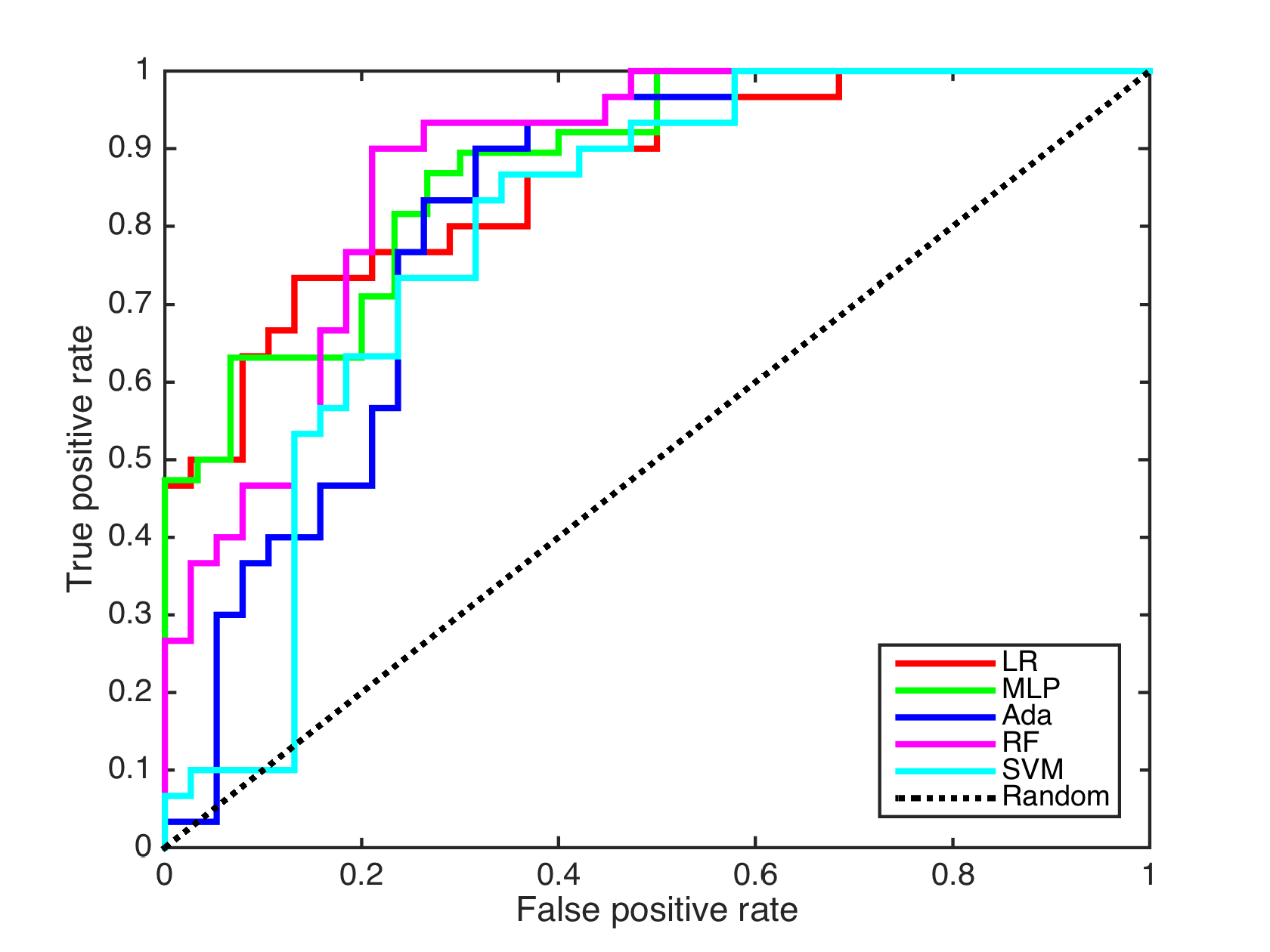}
\label{fig:p1roc}}
\quad
\subfigure[ROC Curve for Raw Data]{
\includegraphics[width=0.47 \linewidth]{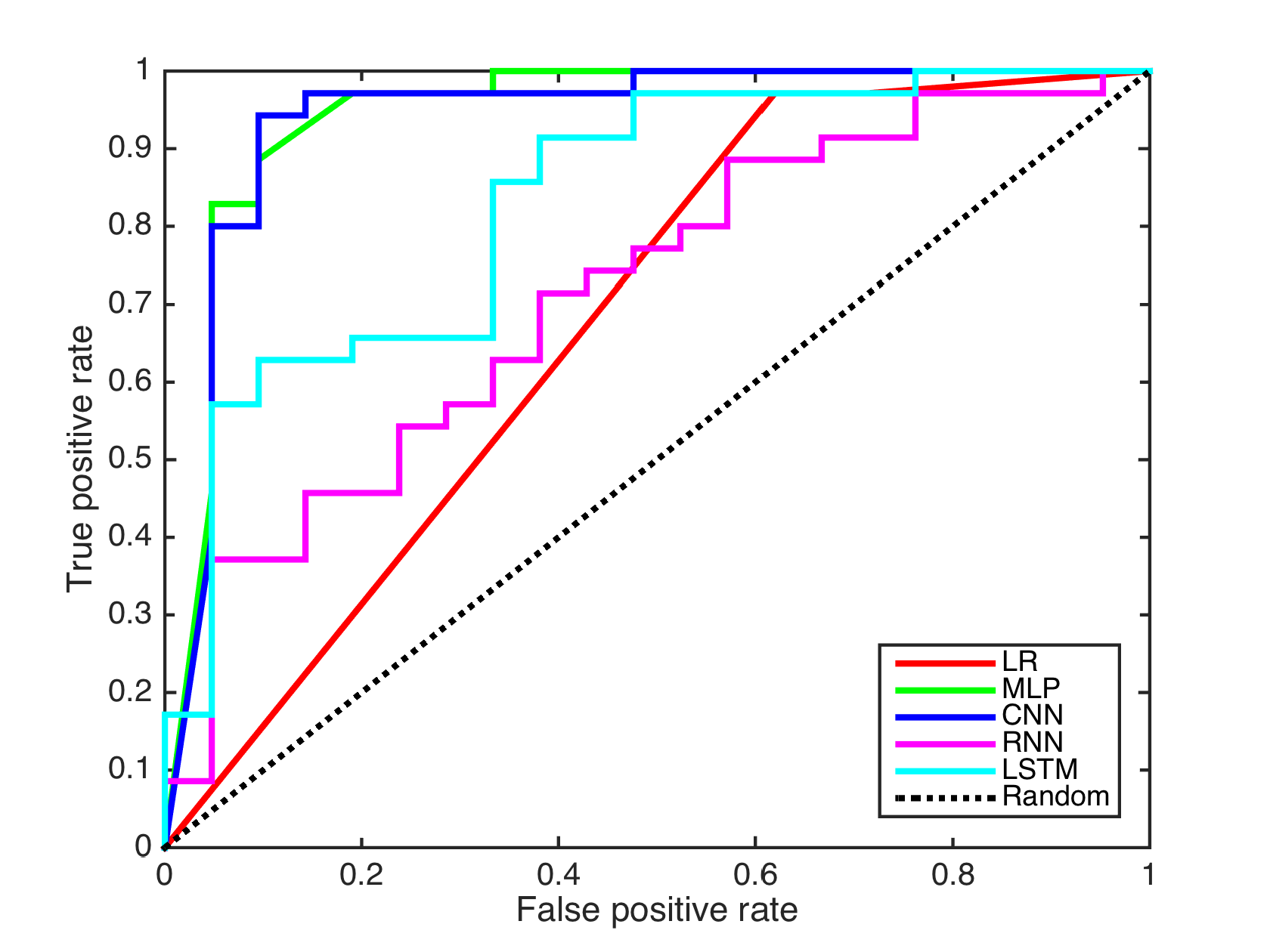}
\label{fig:p2roc}}
\end{figure*}

\begin{table*}
\centering
\caption{Results from RAHAR Output}
	\begin{tabular}{c|cc|cc|cc|cc|cc}
	    \hline
	    &\multicolumn{2}{c}{AU-ROC} & \multicolumn{2}{c}{F1 Score}& \multicolumn{2}{c}{Recall}& \multicolumn{2}{c}{Precision}& \multicolumn{2}{c}{Accuracy}
	    \\
	    \hline
	    &				SE+AL	 &	 RAHAR	 & 	SE+AL 	& 	RAHAR 	&	 SE+AL 	& 	RAHAR 	& 	SE+AL 	& 	RAHAR 	&	 SE+AL 	& 	RAHAR 
	    \\
	    \hline
	    \hline
	    Ada		&	0.7489	&	0.8132	&	0.5574	&	0.6885	&	0.5484	&	0.5526	&	0.5667	&	0.9130	&	0.6966	&	0.7206\\
	    RF 		&	0.8115	&	0.8746	&	0.6885	&	0.7500	&	0.6774	&	0.6316	&	0.7000	&	0.9231	&	0.7865	&	0.7647\\
	    SVM		&	0.7497	&	0.7895	&	0.3721	&	0.7077	&	0.2581	&	0.6053	&	0.6667	&	0.8519	&	0.6966	&	0.7206\\
	    LR 		&	-*	&	0.8649	&	-*		&	0.6875	&	-*		&	0.5789	&	-*		&	0.8462	&	-*		&	0.7059\\
        MLP 	&	-*	&	0.8781	&	-*		&	0.5385	&	-*		&	0.3684	&	-*		&	1.0000	&	-*		&	0.6471\\
	    \hline
        \multicolumn{11}{@{}l}{{\scriptsize * Metrics are not included for models producing uniform classification predictions.}} \\
	\end{tabular}
	\label{tab:RAHAR}
\end{table*}


\subsection{Raw Accelerometer Data Results}
RAHAR is essentially a pre-processing algorithm that creates features for improving the performance of classical methods. With deep learning this is not necessary. High dimensional data requires latent compressed abstract feature representation. To leverage the power of deep learning, we test the raw accelerometer data aggregated into epochs of one minute as input. This eliminates the need for pre-processing, and provides a richer input data set for model building. 

CNN and MLP performed the best with AUC scores showing an improvement from LR by 46\%. 

\subsubsection{Logistic Regression}
For logistic regression, the configuration utilizes mini-batch size and dropout ratio. The optimal settings were with a mini-batch size set to 5, and a dropout ratio of 0.5. 

\subsubsection{Multi Layer Perceptron}
For a multi layer percepton with 1 hidden layer, the size of the hidden layer is also necessaary. The optimal settings were with a mini-batch size of 20, dropout ratio of 0.1 and a higgen layer size of 15.

\subsubsection{Convolutional Neural Network}
Convolution networks instead require the number of filters, the size of the filters and a max pool length, in addition to the mini-batch size and dropout ratio. The best results were obtained with 25, 5, 4, 5, 0.0 respectively.

\subsubsection{Recurrent Neural Network}
An RNN requires the size of the mini-batch, dropout ratio and hidden layer size. It performed best with a mini-batch size of 5, a dropout ratio of 0.1 and a hidden layersize of 75. 

\subsubsection{Long-Short Term Memory Cell}
Lastly, LSTM requires the same as RNNs. A mini-batch size of 5, dropout ratio of 0.5 and a hidden layer size of 100 were optimal. 










\begin{table*}
\centering
\caption{Results from Raw Accelerometer Data}
	\begin{tabular}{c|c|c|c|c|c}
	    \hline
	    &				AU-ROC & F1-Score	 &	 Precision	 & 	Recall & Accuracy
	    \\
	    \hline
	    \hline
	    LR 		&	0.6463 & 	0.8193	&	0.7083	&	0.9714 	& 	0.7321\\
	    MLP 	&	0.9449 &	0.9118	&	0.9394	&	0.8857	& 	0.8929\\
	    CNN		&	0.9456	&	0.9444	&	0.9189	&	0.9714 	& 	0.9286\\
        RNN		&	0.7143	&	0.7711	&	0.6667	&	0.9143 	& 	0.6607\\
        LSTM		&	0.8680	&	0.8831	&	0.8095	&	0.9714 	& 	0.8393\\
	    \hline
	\end{tabular}
	\label{tab:Raw}
\end{table*}

\begin{table*}
\centering
\caption{Sensitivity and Specificity}
	\begin{tabular}{c|c|c}
	    \hline
	    &				Sensitivity & Specificity
	    \\
	    \hline
	    \hline
	    LR 		&	0.9714 	&	0.3333\\
	    MLP 	&	0.8857	&	0.9048\\
	    CNN		&	0.9714	&	0.8571\\
        RNN		&	0.9143	&	0.2381\\
        LSTM		&	0.9714	&	0.6190\\
	    \hline
	\end{tabular}
	\label{tab:SenSpe}
\end{table*}

\section{Discussion}


The objectives of our study were to \Ni evaluate deep learning as a prediction model for sleep quality, following the state-of-the-art methodology, and \Nii to leverage deep learning characterstics to test its overall predictive power. 

Table \ref{tab:RAHAR} shows results on pre-processed, feature-reduced data and illustrates that using deep learning as an alternative to traditional classification models shows a moderate AUC improvement. Although the improvement is small, the success of MLP on a dataset with highly limited features shows its potential to work on smaller, more limited datasets. 

Table \ref{tab:Raw}, shows the results of several deep learning models on the raw accelerometer data. Linear regression was not able to perform as well given the raw data, proving the limitations of classical classification models. MLP performed 7\% better with the expanded feature set, i.e. the raw data versus RAHAR data. Although RNN and LSTM saw an overall improvement to LR, they were limited in their effectivness. The granularity and sequential dependencies of the data lead to a vanishing gradient. Preliminary exploration into aggregating the data into longer duration epochs indicated a potential improvement in both RNN and LSTM. This idea needs further exploration and is beyond the scope of this paper. 

The sensitivity (also known as recall) and specificity of each of the models is reported in Table \ref{tab:SenSpe}. The high sensitivity values of each of the models indicate that deep learning has a strong capability of correctly identifying individuals with good sleep patterns from their preceeding awake activity. The specificity is high for both MLP and CNN, indicating that these models were also able to successfully identify those with poor sleep patterns. 

Overall, CNN performed the best. It scored an AUC of 0.9456, with an F1 Score and Accuracy of 0.9444 and 0.9286, respectivley.

\section{Discussion on Clinical and Health Informatics Importance}

Actigraphy can lead to a paradigm shift in the study of two key lifestyle behaviours (sleep and physical activity), just as ECG (electrocardiography) became crucial for cardiology and clinical research. To achieve that paradigm shift it is necessary to develop new algorithms and tools to analyze this type of data. Furthermore, huge data sets of actigraphy data are emerging from health research, including the study of patients, healthy populations, and epidemiological studies. Furthermore, millions of consumers are buying wearables that incorporate activity sensors. This burst of human activity data is a great opportunity for health research but it requires the development of new tools and approaches. This paper has explored how Deep Learning is one of the most promising technologies for the study of human activity data.

Finally, systematic research on the use of wearables is becoming crucially important with the current boom of sleep apps \cite{behar2013review}. There are serious concerns about the quality of some of those apps \cite{bhat2014there}. Although we can assume that sleep apps relying on activity sensors are becoming more reliable, we cannot be certain since there is scant quantification of this. In this context it becomes more important to promote research in actigraphy data analysis.

\section{Conclusion}
Our results show using a convolutional neural network (CNN) on the raw wearables output improves the predictive value by an additional 8\% compared with the state-of-the-art methodology using classical classification approaches. These positive results in a small data set, show the potential of deep learning for analyzing actigraphy data for sleep research. They also illustrate the potential value of integrating these algorithms into  clinical and research practice.  

Our study shows the feasibility of deep learning in sleep research using actigraphy. This is of paramount importance because deep learning eliminates the need for data pre-processing and simplifies the overall workflow in sleep research.



\ifCLASSOPTIONcaptionsoff
  \newpage
\fi



\bibliographystyle{IEEEtran}
\bibliography{jbhi16_sleep}
\end{document}